\documentclass[10pt,leqno]{amsart}
\usepackage[utf8]{inputenc}
\usepackage{graphicx}
\usepackage{indentfirst,csquotes}
\usepackage{pgfplots}
\usepackage{tcolorbox}
\usepackage{graphicx}
\usepackage{indentfirst,csquotes}
\usepackage{amssymb,amsthm,amsmath}
\usepackage{xcolor,paralist,hyperref,fancyhdr,etoolbox} 
\usepackage{braket}
\usepackage{subcaption} 
\usepackage{booktabs} 
\usepackage{tikz}
\usepackage{standalone}
\usepackage{pgfplots}
\usepackage{lmodern}
\usepackage{import}
\usepackage{algorithm}
\usepackage{algpseudocode}
\usepackage{graphicx}
\usepackage{amssymb,amsmath,mathtools}
\usepackage{hyperref}
\usepackage[T1]{fontenc}
\usepackage{braket}
\usepackage{booktabs}
\usepackage{tcolorbox}   
\usepackage{tikz}        
\usepackage{listings}    
\usepackage{xcolor}      
\usepackage{tikz}
\usetikzlibrary{positioning, arrows.meta, shapes.geometric}

\tikzstyle{block} = [rectangle, draw, text centered, rounded corners, minimum height=2em]
\tikzstyle{line} = [draw, -stealth, thick]
\tikzstyle{cloud} = [ellipse, draw, text centered, minimum height=2em, thick]

\usetikzlibrary{automata, positioning}
\usetikzlibrary{shapes.geometric, arrows, positioning}

\everymath=\expandafter{\the\everymath\displaystyle}

\tikzstyle{block} = [rectangle, draw, text centered, rounded corners, minimum height=2em]
\tikzstyle{line} = [draw, -stealth, thick]
\tikzstyle{cloud} = [ellipse, draw, text centered, minimum height=2em, thick]
\tikzstyle{dashedcloud} = [ellipse, draw, dashed, text centered, minimum height=2em, thick]
\usetikzlibrary{positioning, arrows.meta, shapes.geometric, decorations.pathreplacing}

\tikzstyle{startstop} = [rectangle, rounded corners, minimum width=1.5cm, minimum height=0.5cm,text centered, draw=black, fill=red!30]
\tikzstyle{io} = [trapezium, trapezium left angle=70, trapezium right angle=110, minimum width=1cm, minimum height=0.5cm, text centered, draw=black, fill=blue!30]

\tikzstyle{process} = [rectangle, minimum width=3cm, minimum height=0.5cm, text centered, draw=black, fill=orange!30]
\tikzstyle{decision} = [diamond, minimum width=0.5cm, minimum height=0.1cm, text centered, draw=black, fill=green!30]
\tikzstyle{process2} = [rectangle, minimum width=1cm, minimum height=0.5cm, text centered, draw=black, fill=orange!30]
\tikzstyle{arrow} = [thick,->,>=stealth]
\usetikzlibrary{shapes,shadows,arrows,positioning,angles,quotes}
\tikzset{My Arrow Style/.style={single arrow, fill=black!15, anchor=base, align=center,text width=2.3cm}}
\tikzstyle{arrow} = [thick,->,>=stealth]
\usetikzlibrary{calc,trees,positioning,arrows,fit,shapes,calc}

\newtheorem{theorem}{Theorem}
\newtheorem{definition}[theorem]{Definition}

\hypersetup{
  colorlinks=true,
  linkcolor=blue,    
  citecolor=blue,    
  filecolor=blue,    
  urlcolor=blue      
}

\begin{document}

\title[Verifying and Explaining RL Policies for Platelet Inventory]{COOL-MC: Verifying and Explaining RL Policies for Platelet Inventory Management}
\author[Dennis Gross]{Dennis Gross}
\thanks{Email:\quad\texttt{dennis@artigo.ai}  \quad\textbar\quad \emph{COOL-MC}:\quad \url{https://github.com/LAVA-LAB/COOL-MC}}
\maketitle


\begin{abstract}
Platelets expire within five days. Blood banks face uncertain daily demand and must balance ordering decisions between costly wastage from overstocking and life-threatening shortages from understocking. \emph{Reinforcement learning (RL)} can learn effective ordering policies for this \emph{Markov decision process (MDP)}, but the resulting neural policies remain black boxes, hindering trust and adoption in safety-critical domains.
We apply \emph{COOL-MC}, a tool that combines RL with probabilistic model checking and explainable RL, to verify and explain a trained policy for the MDP on platelet inventory management inspired by Haijema et al. By constructing a policy-induced discrete-time Markov chain (which includes only the reachable states under the trained policy to reduce memory usage), we verify PCTL properties and provide feature-level explanations.
Results show that the trained policy achieves a 2.9\% stockout probability and a 1.1\% inventory-full (potential wastage) probability within a 200-step horizon, primarily attends to the age distribution of inventory rather than other features such as day of week or pending orders.
Action reachability analysis reveals that the policy employs a diverse replenishment strategy, with most order quantities reached quickly while several are never selected.
Counterfactual analysis shows that replacing medium-large orders  with smaller ones leaves both safety probabilities nearly unchanged, indicating that these orders are placed in well-buffered inventory states.
This first formal verification and explanation of an RL platelet inventory management policy demonstrates \emph{COOL-MC}'s value for transparent, auditable decision-making in safety-critical healthcare supply chain domains.
\end{abstract}


\section{Introduction}
Platelets are essential for treating patients with serious conditions such as cancer~\cite{wang2025platelet} and bleeding~\cite{agarwal2024platelet}.
Since platelets are obtained exclusively through voluntary donations and have a very short shelf life of only about five days, managing their inventory is particularly challenging~\cite{rajendran2019inventory}.
Hospitals must repeatedly decide how many units to order from regional blood centers and how to allocate them across patients, all under uncertain future demand~\cite{belien2012supply}.
Each ordering decision directly affects the availability of supply in subsequent periods. Ordering too many units leads to costly wastage of a scarce resource, while ordering too few risks critical shortages that can endanger patients.
This sequential decision-making problem, combined with stochastic demand and strict expiration constraints, makes platelet inventory management a compelling safety-critical application for \emph{reinforcement learning (RL)}~\cite{valizadeh2026platelet,farrington2023deep,farrington2025machine,boute2022deep}.

This platelet inventory management problem can be modeled as a \emph{Markov decision process (MDP)}~\cite{haijema2009blood}, in which an RL agent repeatedly observes the current state of the environment, chooses an action, and receives a penalty signal~\cite{valizadeh2026platelet,maggiar2025structure}.
The state describes the current day of the week, pending orders, and the number of platelet units in stock at each remaining shelf life.
The action determines how many new units to order, and the environment penalizes the agent for each unit that expires unused and for each unit of demand that cannot be fulfilled.
Through repeated interactions, the agent learns a \emph{memoryless policy} that maps each observed state to an order decision, thereby minimizing the total penalty over time.
Modern RL algorithms typically represent this policy using a \emph{neural network (NN)}~\cite{DBLP:journals/corr/MnihKSGAWR13}, enabling them to generalize across large state spaces and outperform traditional optimization methods~\cite{DBLP:conf/setta/GrossJJP22}.

However, RL policies can exhibit \emph{unsafe behavior}~\cite{DBLP:conf/setta/GrossJJP22} such as systematically under-ordering on certain days, leading to dangerous shortage periods~\cite{valizadeh2026platelet}, as penalties are often not enough for complex safety requirements~\cite{DBLP:journals/aamas/VamplewSKRRRHHM22}, and NN-based policies are additionally \emph{hard to interpret} because the internal complexity of NNs hides the reasoning behind each decision~\cite{DBLP:journals/ml/Bekkemoen24}.
In the platelet setting, this opacity is particularly problematic: blood bank managers need to understand \emph{why} a policy orders a certain quantity on a given day and whether it accounts for the age distribution of stock.
Without such understanding, adoption of RL-based ordering policies in practice remains challenging~\cite{boute2022deep}.

To resolve the issues mentioned above, formal verification methods like \emph{probabilistic model checking}~\cite{baier2008principles}, which exhaustively analyzes all possible behaviors of a stochastic system to compute exact probabilities of satisfying formally specified properties, has been proposed to reason about the safety of RL policies~\cite{yuwangPCTL,DBLP:conf/formats/HasanbeigKA20,DBLP:conf/atva/BrazdilCCFKKPU14,DBLP:conf/tacas/HahnPSSTW19} and \emph{explainable RL methods} to interpret trained RL policies~\cite{DBLP:journals/csur/MilaniTVF24,gross2025pctl,DBLP:conf/esann/GrossS24,gross2024enhancing}.

\emph{COOL-MC} addresses both of these challenges by combining RL with probabilistic model checking and explainability~\cite{DBLP:conf/setta/GrossJJP22,gross2025pctl,DBLP:conf/esann/GrossS24,gross2024enhancing}: after learning a policy via RL, it constructs only the reachable state space induced by that policy, yielding a \emph{discrete-time Markov chain (DTMC)} amenable to probabilistic model checking and explainable RL methods~\cite{DBLP:conf/setta/GrossJJP22}.
By constructing only the reachable states under the trained policy rather than enumerating the full state space, this approach mitigates the curse of dimensionality that renders classical methods such as full-MDP model checking intractable as the number of states grows~\cite{DBLP:conf/allerton/KwiatkowskaNP10,DBLP:conf/setta/GrossJJP22}.

On the verification side, PCTL properties formalize safety-relevant questions about the policy's behavior, such as ``Does the policy reach stockout with a probability of less than 5\%?'' or ``Does the policy eventually place a large order?''
Probabilistic model checking then evaluates each of these properties on the induced DTMC, returning both a yes-or-no verdict on whether the property is satisfied and exact quantitative values, such as probabilities and expected steps, that characterize the specified policy behavior~\cite{DBLP:journals/sttt/HenselJKQV22,DBLP:conf/setta/GrossJJP22}.

On the explainability side, feature pruning removes individual state observation features and measures the resulting change in the policy's behavior via probabilistic model checking~\cite{DBLP:conf/esann/GrossS24}, thereby revealing which inventory characteristics drive ordering decisions.
Additionally, state labeling in the induced DTMC via explainability methods such as feature-importance permutation ranking~\cite{breiman2001random} enables identification of which state features most strongly influence the policy's decisions across different ordering trajectories~\cite{gross2025pctl}.
Action labeling annotates each state with the ordering level chosen by the policy, enabling formal queries to detect specific failure modes, for example, whether the policy consistently orders too few units along paths that eventually lead to a stockout.
Beyond explanation, counterfactual analysis enables targeted what-if interventions, such as replacing a specific ordering action with a simpler one or increasing storage capacity, and then re-verifying the model to assess the impact on safety outcomes without retraining the policy.

To date, however, \emph{COOL-MC}'s capabilities have not been applied to the verification and explanation of RL policies for platelet inventory management.

In this paper, we demonstrate that \emph{COOL-MC} can be applied to the platelet inventory management MDP inspired by Haijema et al.~\cite{haijema2009blood} as a case study.
This MDP models the daily ordering planning problem for a regional blood bank in the North-East Netherlands, where the objective is to minimize the combined cost of platelet shortages and full-inventory issues (potential wastage) under stochastic daily demand.
States represent the day of the week, pending orders, and the age-structured inventory, actions correspond to ordering levels, and penalties penalize shortages and wastage with a 5:1 cost ratio.

Using \emph{COOL-MC}, we train a \emph{Proximal Policy Optimization (PPO)} policy~\cite{schulman2017proximal} on the platelet inventory MDP and construct its induced DTMC to verify and explain the learned policy ordering behavior with the approaches mentioned above.

Our analysis reveals properties of the learned policy that are not captured by standard reward-based evaluation.
The trained policy achieves a 2.9\% stockout probability and a 1.1\% inventory-full (potential wastage) probability within a 200-step horizon, primarily attends to the age distribution of inventory rather than other features such as day of week or pending orders.
Feature-importance permutation ranking confirms this picture at the per-state level, showing that inventory age classes dominate the policy's action selection across ordering trajectories.
Action labeling reveals that the policy employs a diverse replenishment strategy, with most order quantities reached quickly while several are never selected.
Counterfactual analysis shows that replacing medium-large orders with smaller ones leaves both safety probabilities nearly unchanged, indicating that these orders are placed in well-buffered inventory states.

\textbf{Our main contribution} is the demonstration of \emph{COOL-MC} as a tool for platelet inventory management applications, combining RL with probabilistic model checking and explainability to provide the first formal analysis of a platelet inventory management RL policy, including verified safety properties, feature-level explanations via pruning and permutation ranking, action-level behavioral profiling, and counterfactual what-if analysis of the trained RL ordering~policy.

\section{Related Work}
\label{sec:related_work}
Our work lies at the intersection of platelet and perishable inventory optimization and safe, explainable RL.

\subsection{Platelet Inventory Optimization}
The foundational work by Haijema et al.~\cite{haijema2009blood} formulated platelet ordering planning as a stochastic dynamic programming and provided an MDP which we use in this study.
Subsequent studies extended this line with stochastic integer programming heuristics~\cite{rajendran2017platelet,rajendran2019inventory} and approximate dynamic programming under endogenous shelf-life uncertainty~\cite{abouee2022data,abouee2025platelet}.

Data-driven learning approaches~\cite{ahmadi2022intelligent,ahmadimanesh2020designing,abbasi2020predicting,zhu2025intelligent,abolghasemi2025machine} and RL~\cite{eisenach2024neural,jiang2009case,maggiar2025structure,dehaybe2024deep,madeka2022deep,giannoccaro2002inventory,boute2022deep,rana2015dynamic,qiao2024distributed,wang2025iot,temizoz2025deep,jullien2022simulation,nomura2023deep,altaf2022deep,farrington2024many,farrington2025going,mohamadi2024application,cheraghi2025dynamic,cheraghi2025learning,valizadeh2026platelet,nomura2025deep,yavuz2024deep,wang2023single,selukar2022inventory,maggiar2025structure} have emerged as a scalable approach for inventory management.
For instance, Kara and Do\u{g}an~\cite{kara2018reinforcement} showed that tabular Q-learning with age-based state representations substantially outperforms stock-quantity-only representations for perishable items with short shelf lives.
Farrington et al.~\cite{farrington2023deep} applied deep RL (PPO and deep Q-Learning) directly to hospital platelet ordering.
In the broader perishable inventory literature, De Moor et al.~\cite{de2022reward} demonstrated that reward shaping stabilizes deep RL training, Gijsbrechts et al.~\cite{gijsbrechts2022can} benchmarked deep RL against state-of-the-art heuristics on canonical inventory problems, and Boute et al.~\cite{boute2022deep} provided a research roadmap that explicitly identifies the \emph{black-box nature} of deep RL policies as a major barrier to adoption.
All of these RL-based approaches evaluate their learned policies exclusively through simulation-based cost metrics.
None of them formally verify the temporal safety properties of the resulting policies, nor do they provide systematic explanations using probabilistic model checking, precisely the gaps our work addresses.

\subsection{Safe and Explainable RL}
Safe reinforcement learning has attracted significant research attention in recent years~\cite{choi2020reinforcement,ganai2024hamilton,landers2023deep,garcia2015comprehensive,dawson2023safe,cheng2019end}.
Ensuring the safety, reliability, and interpretability of RL systems is essential for (blood) platelet inventory management~\cite{abbaspour2021simple,jauhar2025explainable}.
Related work in this area includes shielding methods~\cite{brorholt2024compositional,brorholt2023shielded,corsi2024verification,yang2023safe,konighofer2023online,xiao2023model,konighofer2020shield,jansen2020safe,bloem2015shield,DBLP:conf/aaai/AlshiekhBEKNT18,DBLP:journals/cacm/KonighoferBJJP25,elsayed2021safe} that restrict the agent's actions during training to enforce formal safety specifications~\cite{DBLP:conf/amcc/Bastani21,DBLP:conf/hicss/McCalmonLGCHA23,DBLP:conf/aaai/Carr0JT23,DBLP:conf/ieeecai/QiuJYWZ24,DBLP:journals/cce/GeroldL26}, safe behavioral cloning approaches~\cite{bastani2018verifiable,ashok2019sos,DBLP:conf/tacas/AshokJKWWY21,azeem2025counterexample,azeem20251,DBLP:conf/qest/AshokBCKLT19} that learn policies from verified demonstrations, constrained policy optimization methods~\cite{perkins2002lyapunov,kapoor2020model,hasanbeig2018logically,fulton2019verifiably,fulton2018safe,DBLP:conf/ijcai/WienhoftSSDB023,DBLP:conf/iclr/HogewindSK023,DBLP:conf/aaai/SimaoS023,DBLP:conf/nips/SuilenS0022,DBLP:journals/jair/BadingsRAPPSJ23} that incorporate safety constraints into the learning objective, and verification techniques for analyzing trained RL policies~\cite{ganai2023iterative,fisac2019bridging,fisac2018general,unniyankal2023rmlgym,marzari2025verifying,zolfagharian2024smarla,mannucci2023runtime,lazarus2020runtime,mallozzi2019runtime,jin2022trainify,akintunde2022formal,bacci2020probabilistic,dong2022dependability,bacci2022verified,bacci2021verifying,jin2021learning,tian2023boosting,corsi2021formal,tran2019safety,li2017reinforcement,aksaray2016q,le2024reinforcement,alur2023policy,cai2021reinforcement,hahn2020reward,bozkurt2020control,hasanbeig2019reinforcement,littman2017environment,sadigh2014learning,fu2014probably,wang2024safe,hunt2021verifiably,anderson2020neurosymbolic,junges2016safety,gross2025turn,shao2023sample,jansen2018machine,landers2023deep,DBLP:conf/sigcomm/EliyahuKKS21,DBLP:conf/sigcomm/KazakBKS19,DBLP:journals/corr/DragerFK0U15,DBLP:conf/pldi/ZhuXMJ19,DBLP:conf/seke/JinWZ22,bensalem2024bridging,schmittlearning,venkataraman2020tractable,gangopadhyay2021counterexample,wu2024verified}.
Further safety-related RL policies are presented in~\cite{kwon2024applying}.
However, none of these approaches has been applied to inventory management.
In this work, we apply \emph{COOL-MC} to verify and explain a trained PPO policy for the platelet inventory MDP, providing the first formal safety analysis of an RL-based inventory management policy.

On the explainability side, understanding what drives a trained policy's decisions is critical for trust~\cite{gross2026formallyverifyingexplainingsepsis}.
There exist various explainable RL works in inventory management~\cite{siems2023interpretable,genetti2025evolutionary,teck2025deep}.
For instance, Siems et al. use neural additive models as an intrinsically interpretable policy architecture for inventory RL, where each state feature contributes independently to the ordering decision via learned shape functions~\cite{siems2023interpretable}.
Genetti et al. use evolutionary computation to produce decision-tree policies for supply-chain RL that are inherently interpretable as if-then-else rules~\cite{genetti2025evolutionary}.
Teck et al. apply SHAP as a post-hoc explainability method to a DRL warehouse inventory policy, producing plots that reveal feature contributions to individual decisions~\cite{teck2025deep}.
There is also related work that combines RL with formal methods~\cite{li2019formal,verma2018programmatically}.
What sets our approach apart is that we combine explainability methods with probabilistic model checking to provide temporal explanations via PCTL queries, as introduced in other papers~\cite{DBLP:conf/setta/GrossJJP22,gross2025pctl,DBLP:conf/esann/GrossS24,gross2024enhancing}.

\subsection{Summary.}
Existing RL approaches to platelet inventory learn effective ordering policies but cannot formally guarantee their safety or explain their safe decisions.
Conversely, formal verification methods for RL have been developed for games~\cite{DBLP:conf/setta/GrossJJP22}, job scheduling~\cite{DBLP:conf/aips/GrossS0023}, and clinical treatment~\cite{gross2026formallyverifyingexplainingsepsis}, but have not been applied to inventory problems.
Our work bridges this gap by applying \emph{COOL-MC}'s verification and explainability pipeline to platelet inventory management, providing both PCTL-verified safety guarantees and temporal explanations of the learned RL~policy.

\section{Background}
First, we introduce \emph{Markov decision processes (MDPs)} as the formal framework for sequential decision-making. We then describe probabilistic model checking for verifying policy properties, RL, and different explainability methods for explaining their RL policy decisions.

\subsection{Probabilistic Systems}
A \textit{probability distribution} over a set $X$ is a function $\mu \colon X \rightarrow [0,1]$ with $\sum_{x \in X} \mu(x) = 1$. The set of all distributions on $X$ is $Distr(X)$.

\begin{definition}[MDP]\label{def:mdp}
A \emph{MDP} is a tuple $M = (S,s_0,Act,Tr, rew, AP,L)$
where $S$ is a finite, nonempty set of states; $s_0 \in S$ is an initial state; $Act$ is a finite set of actions; $Tr\colon S \times Act \rightarrow Distr(S)$ is a partial probability transition function and $Tr(s,a,s')$ denotes the probability of transitioning from state $s$ to state $s'$ when action $a$ is taken;
$rew \colon S \times Act \rightarrow \mathbb{R}$ is a reward~function;
$AP$ is a set of atomic propositions;
$L \colon  S \rightarrow 2^{AP}$ is a labeling~function.
\end{definition}

We represent each state $s \in S$ as a vector of $d$ features $(f_1, \dots, f_d)$, where $f_i \in \mathbb{Q}$.
The available actions in $s \in S$ are $Act(s) = \{a \in Act \mid Tr(s,a) \neq \bot\}$ where $Tr(s, a) = \bot$ means that action $a$ is not available in state $s$.
In our setting, we assume that all actions are available at all states.

\begin{definition}[DTMC]\label{def:dtmc}
A \emph{discrete-time Markov chain (DTMC)} is a tuple $D = (S, s_0, Tr, AP, L)$
where $S$, $s_0$, $AP$, and $L$ are as in Definition~\ref{def:mdp}, 
and $Tr \colon S \rightarrow Distr(S)$ is a probability transition~function.
\end{definition}

In many practical settings, the agent does not have direct access to the underlying state $s \in S$ of the MDP. Instead, the agent receives an observation that may represent a transformed view of the true state. We formalize this through an observation function.

\begin{definition}[Observation]
    We define the observation function $\mathbb{O} \colon S \rightarrow O$ as a function that maps a state $s \in S$ to an observation $o \in O$.
    An observation $o \in O$ is a vector of features $(f_1, \dots, f_d)$ where $f_j \in \mathbb{Q}$.
    The observed features may differ from the exact state features.
\end{definition}

A policy operates on observations rather than on the underlying states directly.
In our setting, the environment is fully observable, so the observation function $\mathbb{O}$ maps each state feature directly to its corresponding observation feature, i.e., $\mathbb{O}$ is the identity on the state~features.

\begin{definition}
    A \emph{memoryless deterministic policy $\pi$} for an MDP $M$ is a function $\pi \colon O \rightarrow Act$ that maps an observation $o \in O$ to action $a \in Act$.
\end{definition}

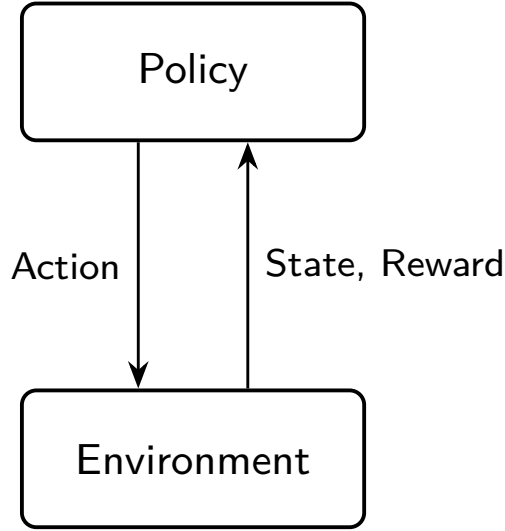
\begin{figure}[]
\centering
\scalebox{0.5}{
    \tikzset{
    process box/.style={
        rectangle,
        rounded corners=3pt,
        draw=black,
        line width=0.8pt,
        fill=white,
        minimum width=2.5cm,
        minimum height=1cm,
        align=center,
        font=\sffamily\small
    },
    arrow/.style={
        -{Stealth[length=2mm, width=1.5mm]},
        line width=0.6pt,
        color=black
    },
    label style/.style={
        font=\sffamily\footnotesize
    }
}

\begin{tikzpicture}

\node[process box] (policy) {Policy};
\node[process box, below=1.8cm of policy] (env) {Environment};

\draw[arrow] ([xshift=-0.4cm]policy.south) -- node[label style, left] {Action} ([xshift=-0.4cm]env.north);

\draw[arrow] ([xshift=0.4cm]env.north) -- node[label style, right] {State, Reward} ([xshift=0.4cm]policy.south);

\end{tikzpicture}
    }
\caption{Sequential decision-making loop. The agent receives an observation and reward from the environment, selects an action according to its policy, and the environment transitions to a new state.}
\label{fig:rl}
\end{figure}

Applying a policy $\pi$ to an MDP $M$ with observation function $\mathbb{O}$ yields an \emph{induced DTMC} $D^\pi$ where all non-determinism is resolved: for each state $s$, the transition function becomes $Tr(s, s') = Tr(s, \pi(\mathbb{O}(s)), s')$.
The induced DTMC fully characterizes the observable behavior of the policy: the states visited, the transitions taken, and the probabilities of all outcomes.
The interaction between the policy and environment is depicted in Figure~\ref{fig:rl}.

\subsection{Probabilistic Model Checking}

\emph{Probabilistic model checking} enables the verification of quantitative properties of stochastic systems.
\emph{COOL-MC} uses \emph{Storm}~\cite{DBLP:journals/sttt/HenselJKQV22} internally as its model checker, which can verify properties of both MDPs and DTMCs.
Among the most fundamental properties are \emph{reachability} queries, which assess the probability of a system reaching a particular state.
For example, one might say: ``The reachability probability of reaching an unsafe state is~0.1.''

\begin{figure}[htbp]
    \centering
    \scalebox{0.5}{
    \tikzset{
    ellipse node/.style={
        ellipse,
        draw=black,
        line width=0.8pt,
        fill=white,
        minimum width=2.2cm,
        minimum height=0.8cm,
        align=center,
        font=\sffamily\footnotesize
    },
    process block/.style={
        rectangle,
        rounded corners=3pt,
        draw=black,
        line width=0.8pt,
        fill=gray!15,
        minimum width=2.4cm,
        minimum height=0.9cm,
        align=center,
        font=\sffamily\small
    },
    arrow/.style={
        -{Stealth[length=2mm, width=1.5mm]},
        line width=0.6pt,
        color=black
    },
    label style/.style={
        font=\sffamily\scriptsize
    }
}

\begin{tikzpicture}[node distance=0.7cm]

\node[ellipse node] (system) {system};
\node[ellipse node, below=0.6cm of system] (systemdesc) {system description};
\node[ellipse node, below=0.6cm of systemdesc] (model) {model};

\node[ellipse node, right=1.2cm of systemdesc] (requirements) {requirements};
\node[ellipse node, below=0.6cm of requirements] (properties) {properties};

\node[process block, below=0.6cm of model, xshift=1.2cm] (modelchecking) {model checking};

\node[ellipse node, below left=0.6cm and 0.2cm of modelchecking] (satisfied) {satisfied};
\node[ellipse node, below right=0.6cm and 0.2cm of modelchecking] (violated) {violated};

\draw[arrow] (system) -- node[label style, right] {modeling} (systemdesc);
\draw[arrow] (systemdesc) -- node[label style, right] {translates to} (model);
\draw[arrow] (requirements) -- node[label style, right] {formalizing} (properties);
\draw[arrow] (model) -- (modelchecking);
\draw[arrow] (properties) -- (modelchecking);
\draw[arrow] (modelchecking) -- (satisfied);
\draw[arrow] (modelchecking) -- (violated);

\end{tikzpicture}
    }
    \caption{General model checking workflow~\cite{DBLP:journals/sttt/HenselJKQV22}. The system is formally modeled, the requirements are formalized, and both are input to a model checker such as \emph{Storm}, which verifies the property.}
    \label{fig:model_checking}
\end{figure}
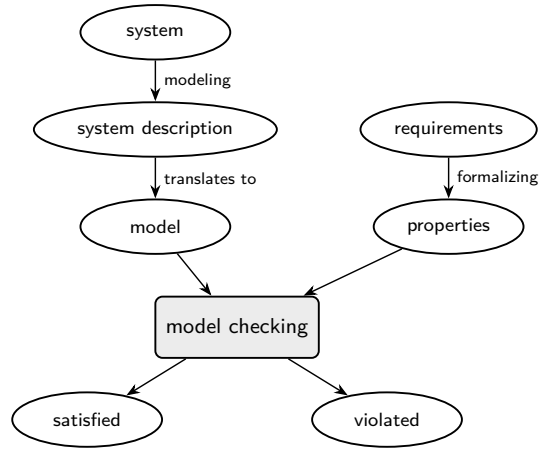

The general workflow for model checking is as follows (see also Figure~\ref{fig:model_checking}).
First, the system is formally modeled using a language such as \emph{PRISM}~\cite{prism_manual}.
Next, the property of interest is formalized in a temporal logic.
Using these inputs, the model checker verifies whether the property holds or computes the relevant probability.

The \emph{PRISM} language~\cite{prism_manual} describes probabilistic systems as a collection of \emph{modules}, each containing typed \emph{variables} and guarded \emph{commands}.
A state of the system is a valuation of all variables, and the state space $S$ is the set of all possible valuations.
Each command takes the~form
\[
\texttt{[action]}\ g \rightarrow \lambda_1 : u_1 + \cdots + \lambda_n : u_n
\]
where $g$ is a Boolean guard over the variables, each $\lambda_j$ is a probability, and each $u_j$ is an update that assigns new values to the variables.
When the system is in a state satisfying $g$, the command can be executed: update $u_j$ is applied with probability $\lambda_j$, transitioning the system to a new state.
For an MDP, each enabled command in a state corresponds to a distinct probabilistic choice, and the nondeterministic selection among enabled commands is resolved by a policy.
Listing~\ref{lst:prism_example} shows a simplified MDP in \emph{PRISM}.
The model declares a single module with a state variable \texttt{s} ranging over integer identifiers.
Each command specifies a transition: a guard on the current state, an action label, and a probabilistic update.
For instance, from state $s=0$, taking action \texttt{a0} leads to state $s=1$ with probability $0.7$ and to state $s=2$ with probability $0.3$.
States $s=2$ and $s=3$ are absorbing terminal states corresponding to different labels.
A reward structure assigns a reward of $+1$ to the $s=2$ state.
Additionally, \emph{PRISM} supports \emph{atomic labels} that annotate states with propositions.
These labels are defined using \texttt{label} declarations, which assign a name to a Boolean condition over the state variables.
For example, \texttt{label "goal" = s=2;} marks all states where $s=2$ with the proposition \textit{goal}.
Such labels serve as the atomic propositions referenced in PCTL properties.
\emph{COOL-MC} allows automated user-specified state labeling of \emph{PRISM} models.
For a complete formal treatment of the \emph{PRISM} semantics, we refer to~\cite{prism_manual}.

\begin{figure}[htbp]
\begin{lstlisting}[
    language={},
    basicstyle=\ttfamily\small,
    keywordstyle=\bfseries,
    commentstyle=\itshape\color{gray},
    frame=single,
    numbers=left,
    numberstyle=\tiny,
    caption={Simplified PRISM model. States are encoded as integers; transitions are guarded by the current state and labeled with actions. Atomic labels annotate states with propositions used in PCTL~properties.},
    label={lst:prism_example},
    escapeinside={(*}{*)}
]
mdp

module example
  s : [0..3] init 0; // state variable

  // From state 0, action a0:
  //   -> state 1 (w.p. 0.7), state 2 (w.p. 0.3)
  [a0] s=0 -> 0.7:(s'=1) + 0.3:(s'=2);

  // From state 0, action a1:
  //   -> state 1 (w.p. 0.4), state 3 (w.p. 0.6)
  [a1] s=0 -> 0.4:(s'=1) + 0.6:(s'=3);

  // From state 1, action a0:
  //   -> state 2 (w.p. 0.8), state 3 (w.p. 0.2)
  [a0] s=1 -> 0.8:(s'=2) + 0.2:(s'=3);

  // Absorbing states (survival and death)
  [end] s=2 -> 1.0:(s'=2); // goal
  [end] s=3 -> 1.0:(s'=3); // empty

endmodule

// Atomic labels for state propositions
label "goal" = s=2;
label "empty" = s=3;

// Reward: +1 for reaching goal state
rewards
  s=2 : 1;
endrewards
\end{lstlisting}
\end{figure}

Properties are specified using PCTL~\cite{hansson1994logic}, a branching-time temporal logic for reasoning about probabilities over paths.
In this paper, we use two path operators.
The \emph{eventually} operator $\lozenge\,\varphi$ (also written $F\,\varphi$) states that $\varphi$ holds at some future state along a path.
The \emph{until} operator $\varphi_1 \;\mathcal{U}\; \varphi_2$ states that $\varphi_1$ holds at every state along a path until a state is reached where $\varphi_2$ holds.
A bounded variant $\lozenge^{\leq B}\,\varphi$ restricts the eventually operator to paths of at most $B$ steps.
A PCTL property has the form
\[
P_{\sim p} (\psi)
\]
where $\psi$ is a path formula (such as $\lozenge\,\varphi$ or $\varphi_1 \;\mathcal{U}\; \varphi_2$), $\sim$ is a comparison operator ($<$, $\leq$, $\geq$, $>$), and $p \in [0,1]$ is a probability threshold.
Beyond checking whether a property is satisfied, \emph{Storm} can compute the exact probability, denoted $P_{=?}(\psi)$.
For example, $P_{=?}(\lozenge^{\leq 200}\;\textit{empty})$ computes the exact probability of reaching a stockout within 200~steps.

In addition to probability queries, PCTL supports \emph{expected time} queries of the form $T_{=?}(F\;\varphi)$, which compute the expected number of steps to first reach a state satisfying $\varphi$.
For instance, $T_{=?}(F\;\texttt{"pr}_{14}\texttt{"})$ returns the expected number of steps until the policy first selects an order of 14~units.
Such queries enable temporal characterization of the policy's ordering behavior, revealing how quickly different actions or inventory configurations are encountered during system evolution.
We refer to~\cite{hansson1994logic,baier2008principles} for a detailed description and additional PCTL operators.

\subsection{Reinforcement Learning}
The standard goal of RL is to learn a policy $\pi$ for an MDP that maximizes the expected accumulated discounted reward~\cite{DBLP:journals/ml/Bekkemoen24}
\[
\mathbb{E}\left[\sum_{t=0}^{N} \gamma^t \cdot rew(s_t, a_t)\right],
\]
where $\gamma \in [0,1]$ is the discount factor, $rew(s_t, a_t)$ is the reward at time step $t$, and $N$ is the episode length.

\subsection{Explainability Methods}\label{sec:xrl}
Explainability methods aim to make trained RL policies understandable~\cite{DBLP:journals/csur/MilaniTVF24}.
Global methods interpret the overall policy's behavior~\cite{DBLP:conf/aiide/SieusahaiG21}, while local methods explain a policy's decision-making in individual states.
In this work, we use four complementary methods: \emph{feature pruning}, \emph{feature permutation importance}, \emph{action labeling}, and \emph{counterfactual action replacement}, each combined with PCTL model checking on the induced~DTMC.

Consider a neural network policy operating on observations with $d$ input features and $|Act|$ output actions, encoding a function $g \colon \mathbb{Q}^d \to \mathbb{R}^{|Act|}$.
The function $g$ is parameterized by a sequence of weight matrices $\vec{W}^{(1)}, \dots, \vec{W}^{(k)}$ with $\vec{W}^{(\ell)} \in \mathbb{R}^{d_\ell \times d_{\ell-1}}$ for $\ell = 1, \dots, k$.
Feature pruning eliminates all outgoing connections from a particular input neuron by setting $\vec{W}^{(1)}_{j,i} = 0$ for all $j$, effectively removing the $i$-th input feature $f_i$ from the policy's decision-making~\cite{DBLP:conf/esann/GrossS24}.
By measuring the resulting change in the policy's behavior (e.g., degradation in stockout probability on the induced DTMC), we quantify the importance of each feature.

Feature permutation importance~\cite{breiman2001random} offers a complementary, model-agnostic measure of feature relevance.
For a given observation $o = \mathbb{O}(s)$ with features $(f_1, \dots, f_d)$, the values of a single feature $f_i$ are randomly permuted across observations while all other features remain fixed.
The policy is then re-evaluated on the permuted inputs, and the change in action selection is measured.
A large change indicates that the policy's decisions depend strongly on $f_i$.
Unlike feature pruning, which permanently modifies the network weights, permutation importance leaves the policy intact. It can be computed per state, providing a local importance ranking that can be used to label states in the induced DTMC for further PCTL analysis.

Action labeling annotates each state in the induced DTMC with the action selected by the trained policy~\cite{gross2026formallyverifyingexplainingsepsis}.
These labels become atomic propositions that can be referenced in PCTL properties, enabling formal queries over the policy's decision behavior, such as detecting whether certain ordering levels are consistently chosen before reaching undesirable states.

Counterfactual action replacement modifies the policy by substituting one action with another across all states where it is selected, and then reconstructs and re-verifies the induced DTMC~\cite{gross2024enhancing}.
This allows targeted what-if analysis that isolates the effect of specific ordering decisions on safety outcomes. For example, one can assess how the probabilities of stockout and wastage change if all medium- and large-order quantities are replaced with smaller ones, without retraining the RL policy.

\section{Methodology}
Our methodology proceeds in four stages.
First, we encode the platelet inventory management MDP~\cite{haijema2009blood} in the \emph{PRISM} modeling language, defining states, actions, transition dynamics, and penalty structures that capture the daily ordering planning problem (\S\ref{subsec:mdp-encoding}).
Second, we train a deep RL policy using PPO to learn an ordering strategy that minimizes the combined cost of shortages and outdating (inventory-full) (\S\ref{subsec:rl-training}).
Third, we verify the trained policy on its induced DTMC by applying PCTL queries to assess safety properties such as stockout probability and to characterize ordering trajectories (\S\ref{subsec:policy-verification}).
Fourth, we explain the policy's decision-making through feature pruning, feature-importance permutation ranking, action labeling, and counterfactual action replacement, combined with PCTL queries on the induced DTMC (\S\ref{subsec:policy-explanation}).
Finally, we describe the limitations of our methodology (\S\ref{subsec:limitations}).

\subsection{MDP Encoding}\label{subsec:mdp-encoding}

We encode the platelet inventory MDP inspired by Haijema et al.~\cite{haijema2009blood} in the \emph{PRISM} modeling language. This MDP models the daily ordering planning problem for a regional blood bank in the North-East Netherlands, with the objective of minimizing the combined cost of platelet shortages and outdating.
The model parameters are derived from operational data of approximately 144 platelet pools demanded per week~\cite{haijema2009blood}, following the aggregation and downsizing procedure described therein.

Our \emph{PRISM} encoding uses a two-phase structure per day that differs from Haijema et al.'s original formulation: in phase~0, the controller makes a nondeterministic ordering decision; in phase~1, stochastic demand is realized and inventory is updated. Haijema et al.\ model each day as a single decision epoch in which the production decision and demand realization occur implicitly within one transition. Our two-phase separation explicitly encodes this interleaving of controller actions and nature's moves within \emph{PRISM}'s modeling framework, introducing the additional state variables $\mathit{pend}$ and $\mathit{ph}$ (see below).
The MDP cycles indefinitely through the weekly schedule, 
matching the infinite-horizon formulation of Haijema et al.

\subsubsection{State space}
The state is represented by 8 features: the day of the week $d \in \{0, \ldots, 6\}$ (Monday through Sunday), five age-structured inventory features $x_1, \ldots, x_5 \in \{0, \ldots, S_{\max}\}$ where $x_r$ denotes the number of aggregated units with $r$ days of residual shelf life, a pending order feature $\mathit{pend} \in \{0, \ldots, S_{\max}\}$ representing units ordered today that become available the next morning (this feature is part of our \emph{PRISM} encoding to explicitly track the one-day lead time; it is not a separate state feature in Haijema et al.'s original formulation), and a phase feature $\mathit{ph} \in \{0, 1\}$.
The total available inventory is $\mathit{tinv} = \sum_{r=1}^{5} x_r$. Following the downsizing procedure of Haijema et al., one aggregated unit corresponds to 4 real platelet pools, and $S_{\max} = 30$.

\subsubsection{Actions}
The action set consists of 31 ordering levels $\{\mathit{pr}_0, \mathit{pr}_1, \ldots, \mathit{pr}_{30}\}$, where action $\mathit{pr}_k$ orders $k$ aggregated units. On weekdays (Monday--Friday), all actions satisfying $\mathit{tinv} + k \leq S_{\max}$ are available; on weekends, only $\mathit{pr}_0$ (no ordering) is enabled.
The chosen order amount is stored in \textit{pend} and is \emph{not} available to serve same-day demand, matching the specification of Haijema et al.\ that ordering is released at the end of the day and becomes available the next morning~\cite{haijema2009blood}.
In phase~1, the environment executes a single demand-realization action $\mathit{dm}$.

\subsubsection{Demand and transition dynamics}
Demand on day $d$ follows a Poisson distribution with day-dependent rate $\lambda_d$, truncated to the range $\{0, \ldots, K_{\max}\}$ with $K_{\max} = 20$ and renormalized. The daily rates (in aggregated units) are $\lambda_{\text{Mon}} = 6.5$, $\lambda_{\text{Tue}} = 5.0$, $\lambda_{\text{Wed}} = 8.0$, $\lambda_{\text{Thu}} = 5.0$, $\lambda_{\text{Fri}} = 6.5$, $\lambda_{\text{Sat}} = 1.75$, and $\lambda_{\text{Sun}} = 3.25$, corresponding to the weekly demand pattern of $[26, 20, 32, 20, 26, 7, 13]$ pools divided by the aggregation factor of~4.
The Poisson probabilities are computed inline using \emph{PRISM} formulas via unnormalized weights $w_k = \lambda_d^k / k!$ and normalization $p_k = w_k / \sum_{j=0}^{K_{\max}} w_j$, ensuring probabilities sum to exactly~1 for any parameter values.

Given demand realization $b$, the transition encodes three simultaneous operations:
\begin{enumerate}
    \item \textbf{FIFO issuing:} Demand is served oldest-first. The remaining demand after consuming age classes $1$ through $r$ is $\max(0,\, b - \sum_{i=1}^{r} x_i)$, and the leftover stock at age $r$ is $\max(0,\, x_r - \max(0,\, b - \sum_{i=1}^{r-1} x_i))$.
    \item \textbf{Outdating and aging:} Any leftover $x_1$ (oldest stock) is discarded. The remaining stock shifts down one age class: $x_r' = \text{leftover}_{r+1}$ for $r = 1, \ldots, 4$.
    \item \textbf{Delivery:} Today's pending order arrives as fresh stock: $x_5' = \mathit{pend}$.
\end{enumerate}
The day then advances ($d' = (d+1) \bmod 7$) and the phase resets to~0.

\subsubsection{Penalty structure}
Penalties are defined as state-based expected costs at phase~1, computed over the demand distribution using formulas:
\begin{align*}
    \mathbb{E}[\text{shortage}] &= \textstyle\sum_{k=0}^{K_{\max}} p_k \cdot \max(0,\, k - \mathit{tinv}), \\
    \mathbb{E}[\text{outdating}] &= \textstyle\sum_{k=0}^{K_{\max}} p_k \cdot \max(0,\, x_1 - k).
\end{align*}
The combined cost reward is $C_S \cdot \mathbb{E}[\text{shortage}] + C_O \cdot \mathbb{E}[\text{outdating}]$, where $C_S = 5$ and $C_O = 1$ reflect the 5:1 shortage-to-outdating cost ratio of Haijema et al.

\subsubsection{Labels}
States are labeled with atomic propositions to enable PCTL property specification such as \textit{empty} ($\mathit{tinv} = 0$), \textit{full} ($\mathit{tinv} = S_{\max}$), and \textit{weekend} ($d \geq 5$).


\subsubsection{Summary} The encoded \emph{PRISM} MDP is formally defined as
\begin{gather*}
    S = \{0,\ldots,6\} \times \{0,\ldots,S_{\max}\}^5 \times \{0,\ldots,S_{\max}\} \times \{0, 1\} \\
    \mathit{Act} = \{\mathit{pr}_0, \mathit{pr}_1, \ldots, \mathit{pr}_{30}\} \cup \{\mathit{dm}\} \\
    \mathit{rew}(s) = C_S \cdot \mathbb{E}[\text{shortage} \mid s] + C_O \cdot \mathbb{E}[\text{outdating} \mid s] \quad \text{if } \mathit{ph} = 1.
\end{gather*}

\subsection{RL Policy Training}\label{subsec:rl-training}

We train a deep RL policy $\pi$ on the platelet inventory MDP using PPO.
The policy network is a feedforward neural network with three fully connected hidden layers of 256 neurons.
The agent is trained for 25{,}000 episodes with a learning rate of $\alpha = 3 \times 10^{-4}$, discount factor $\gamma = 0.99$, batch size 64, and a maximum episode length of 200~steps.

\subsection{Policy Verification}\label{subsec:policy-verification}

For the trained policy $\pi$, \emph{COOL-MC} constructs the induced DTMC $D^\pi$ by exploring only the reachable state space under $\pi$ (see Algorithm~\ref{alg:qverifier}).
Starting from the initial state $s_0$, the construction proceeds as a depth-first traversal.
At each visited state $s$, the policy selects an action $a = \pi(s)$, and all successor states $s'$ with $Tr(s, a, s') > 0$ are added to the induced DTMC along with their transition probabilities.
The procedure recurses into each successor that has not yet been visited and is relevant for the property under verification.
This incremental approach offers two key advantages.
First, it constructs only the fragment of the full MDP that is actually reachable under $\pi$, which can be substantially smaller than the complete state space, mitigating the state explosion problem.
Second, it creates an induced DTMC rather than an MDP, since all nondeterminism is resolved by $\pi$, reducing the number of states and transitions.

\begin{algorithm}[t]
\caption{\emph{COOL-MC}: Formal Verification of Policies}
\label{alg:qverifier}
\begin{algorithmic}[1]
\Require MDP $M$, trained policy $\pi$, PCTL property $\varphi$, \textsc{Label}(s)
\Ensure Satisfaction result and probability $p$

\Statex
\Statex \textbf{Stage 1: Induced DTMC Construction}
\State $D^\pi \gets (S^\pi, s_0, Tr^\pi, AP, L)$ where $S^\pi \gets \emptyset$, $Tr^\pi \gets \emptyset$
\State \textsc{BuildDTMC}($s_0$)

\Statex
\Statex \textbf{Stage 2: Probabilistic Model Checking}
\State $(result, p) \gets \emph{Storm}.\text{verify}(D^{\pi}, \varphi)$
\State \Return $(result, p)$

\Statex
\Procedure{BuildDTMC}{$s$}
    \If{$s \in S^\pi$ \textbf{or} $s$ is not relevant for $\varphi$}
        \State \Return
    \EndIf
    \State $S^\pi \gets S^\pi \cup \{s\}$
    \State $L^\pi(s) \gets \textsc{Label}(s)$
    \State $a \gets \pi(s)$
    \ForAll{$s' \in S$ where $Tr(s, a, s') > 0$}
        \State $Tr^\pi(s, s') \gets Tr(s, a, s')$
        \State \textsc{BuildDTMC}($s'$)
    \EndFor
\EndProcedure

\end{algorithmic}
\end{algorithm}

Once the induced DTMC is fully constructed, it is passed to \emph{Storm}~\cite{DBLP:journals/sttt/HenselJKQV22} for probabilistic model checking.
We evaluate PCTL properties and queries that formalize safety-relevant questions about the policy's ordering behavior.
For instance, the property $P_{\leq 0.05}(\lozenge^{\leq 200}\;\textit{empty})$ checks whether the probability of reaching a complete stockout within 200~steps remains below 5\%, while the query $P_{=?}(\lozenge^{\leq 200}\;\textit{empty})$ computes the exact probability of that event.

During DTMC construction, states are also annotated with user-specified atomic propositions via the labeling function \textsc{Label}(s).
The \emph{PRISM} model already defines domain-relevant labels (\textit{empty}, \textit{full}, \textit{monday}, \textit{weekend}), and \emph{COOL-MC} extends these with additional labels derived from the policy's behavior and RL explainability analyses, as described in the following~section.

\subsection{Policy Explanation}\label{subsec:policy-explanation}
Verification alone establishes what the policy does (e.g., the probability of reaching stockout) but not \emph{why} it makes particular ordering decisions.
To address this, we apply four explainability methods: feature pruning, feature-importance permutation ranking, action labeling, and counterfactual action replacement.

\subsubsection{Feature pruning}
Feature pruning~\cite{DBLP:conf/esann/GrossS24} provides a global sensitivity analysis of the trained neural network policy.
For each input feature $f_i \in \{d, x_1, x_2, x_3, \dots \}$, all outgoing connections from the corresponding input neuron are set to zero, effectively removing $f_i$ from the policy's decision-making.
The induced DTMC is then reconstructed under the pruned policy, and the PCTL query of interest (e.g., $P_{=?}(\lozenge^{\leq 200}\;\textit{empty})$) is re-computed.
The change in probability relative to the unpruned baseline quantifies the extent to which the policy relies on each feature for shortage avoidance.
A large increase in stockout probability upon pruning a feature indicates that the policy critically depends on that feature; a negligible change suggests redundancy.
This analysis reveals, for instance, whether removing the weekday
$d$ causes the policy to fail to build stock on specific days, such as Fridays.

\subsubsection{Feature-importance permutation ranking}
Feature permutation importance~\cite{breiman2001random} complements pruning with a local, per-state measure of feature relevance.
For each state observation $o$ in the induced DTMC, the values of a single feature $f_i$ are randomly permuted across observations while all other features remain fixed, and the policy is re-evaluated on the permuted inputs.
A large change in action selection indicates that the policy's decision in state $s$ depends strongly on $f_i$.
Each state is labeled with the feature whose permutation most frequently changes the policy's action (e.g., \texttt{imp\_d}, \texttt{imp\_x1}, \ldots, \texttt{imp\_ph}).
These labels are injected into the induced DTMC, enabling PCTL queries such as $P_{=?}(\lozenge^{\leq 10}\;\texttt{imp\_x1})$, which computes the probability of reaching a state where the oldest inventory class is the most important feature, and combined queries such as $P_{=?}(\lozenge\;(\texttt{friday} \wedge \neg\texttt{imp\_d}))$, which detects whether the policy ignores the day-of-week on the critical pre-weekend order~day.

\subsubsection{Action labeling}
Action labeling annotates each state in the induced DTMC with the action selected by the trained policy~\cite{gross2026formallyverifyingexplainingsepsis}, using the notation $\mathit{pr}_k$ for the order of $k$ aggregated units (where $\mathit{pr}_0$ denotes no order and $\mathit{pr}_{30}$ denotes maximum order).
These labels become atomic propositions that can be referenced in PCTL, enabling formal queries directly over the policy's ordering behavior.
For example, $P_{=?}(\lozenge^{\leq 200}\;\mathit{pr}_{30})$ quantifies the probability that the policy resorts to maximum ordering within 200~steps, while $P_{=?}(\lozenge\;(\mathit{pr}_0 \wedge \texttt{monday}))$ detects the reachability probability of no ordering on Mondays.

\subsubsection{Counterfactual action replacement}
Counterfactual action replacement modifies the policy by substituting one action with another across all states where it is selected, and then reconstructs and re-verifies the induced DTMC~\cite{gross2024enhancing}.
This allows targeted what-if analysis that isolates the effect of specific ordering decisions on safety outcomes without retraining the policy.
For example, one can assess how stockout and wastage probabilities change if all medium-large orders ($\mathit{pr}_{14}$, ordering 14~units) are replaced with smaller orders ($\mathit{pr}_{6}$, ordering 6~units).
By re-verifying both $P_{=?}(\lozenge^{\leq 200}\;\textit{empty})$ and $P_{=?}(\lozenge^{\leq 200}\;\textit{full})$ on the counterfactual DTMC, we quantify the trade-off between shortage risk and wastage risk induced by the action change.

\subsection{Limitations}\label{subsec:limitations}

Several limitations should be noted.
First, the platelet inventory MDP uses aggregated demand data from a single regional blood bank in the Netherlands~\cite{haijema2009blood}.
The demand patterns, Poisson rates, and cost parameters are specific to this setting; policies verified on this MDP may not directly transfer to other hospitals or regions with different demand profiles.
Second, \emph{COOL-MC} requires a discrete action space, which is naturally satisfied by the platelet MDP's integer ordering levels.
However, the state space grows combinatorially with $S_{\max}$ and the number of age classes.
While the induced DTMC mitigates this by constructing only the reachable states under a single policy, sufficiently large inventory capacities or finer shelf-life discretizations could still pose scalability challenges.
Third, the \emph{PRISM} model's Poisson demand is truncated at $K_{\max} = 20$ and renormalized, which slightly distorts the tail of the demand distribution.

\section{Experiments}
In this section, we apply the methodology described above to the platelet inventory MDP.
We first describe the experimental setup, including the hardware environment and software tools used.
We then present our analysis in two parts: Part~A covers the training of the PPO agent and the construction and baseline verification of the induced DTMC, establishing the policy's stockout and inventory-full probabilities.
Part~B provides a detailed behavioral analysis of the trained policy, progressively applying permutation importance, feature pruning, action labeling, combined labeling, parameter sensitivity, and counterfactual action replacement to build a comprehensive picture of the policy's ordering behavior and its driving factors.
The experiments are intended to illustrate the range of analyses that \emph{COOL-MC} enables for inventory management settings.

\subsection{Setup}
We executed our experiments in a Docker container with 16\,GB RAM and an AMD Ryzen 7 7735HS processor, running Ubuntu 20.04.5 LTS.
For model checking, we use \emph{Storm} 1.7.0.
Implementation details are provided in the accompanying source code at \url{https://github.com/LAVA-LAB/COOL-MC/tree/sepsis}, which hosts both this and the sepsis case study as joint COOL-MC application examples.

\subsection{Analysis}
We organize the analysis into two parts.
Part~A establishes the baseline by training the PPO policy and verifying its performance and satisfaction of complex safety requirements.
Part~B then systematically investigates the policy's behavior using the explainability and probabilistic model checking introduced in \S\ref{subsec:policy-explanation}.

\subsubsection{Part A: Training and baseline safety}
\begin{figure}
    \centering
    \scalebox{0.5}{
        \input{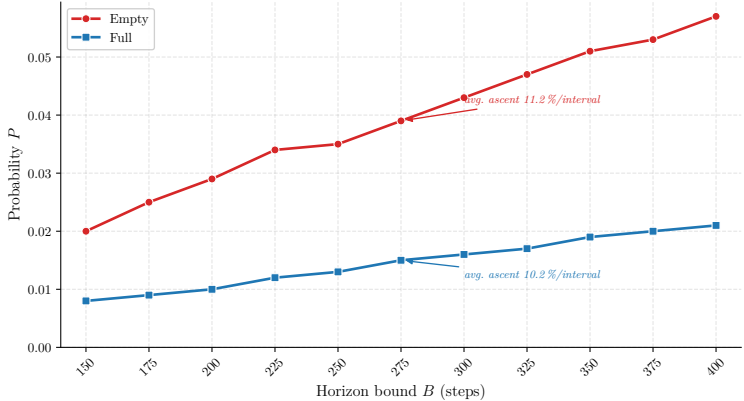}
    }
    \caption{Bounded reachability probabilities
$P_{=?}(\lozenge^{\leq B}\;\textit{empty})$ and
$P_{=?}(\lozenge^{\leq B}\;\textit{full})$ under the trained PPO policy
for increasing horizon bounds $B \in \{150, 175, \dots, 400\}$.
Both probabilities grow monotonically with the horizon, with stockout
risk consistently exceeding inventory-full risk.
The average relative ascent per interval is similar for both properties
(${\approx}\,11.2\%$ for \textit{empty}, ${\approx}\,10.2\%$ for
\textit{full}), indicating that neither risk dominates in its sensitivity
to the planning horizon.}
    \label{fig:different_bounds}
\end{figure}

The PPO agent was trained for 25{,}000 episodes on the platelet
inventory MDP with a maximal step number per episode of 200, achieving a
best sliding-window average reward of $-47.13$.
After training, we constructed the induced DTMCs under the trained policy
for each baseline property. The stockout DTMC comprises 5{,}209
reachable states and 41{,}987 transitions, while the inventory-full DTMC
comprises 5{,}115 states and 40{,}770 transitions, with the slight
difference reflecting the states relevant to each reachability query.
The baseline properties yield
\begin{align*}
P_{=?}(\lozenge^{\leq 200}\;\textit{empty}) &= 0.029, \\
P_{=?}(\lozenge^{\leq 200}\;\textit{full}) &= 0.0105,
\end{align*}
indicating a 2.9\% probability of reaching complete stockout and a
1.1\% probability of reaching a full inventory within 200 steps under
the trained policy.
In Figure~\ref{fig:different_bounds}, we observe the trained policy
performance for different time horizons (max steps).

Both induced DTMCs for 200 time steps are substantially smaller than
the full MDP, which comprises 1{,}684{,}220 states and 20{,}369{,}593
transitions, representing a reduction of over 99.6\% in state space
size. This demonstrates the practical advantage of constructing only the
reachable states under the trained policy, enabling formal verification
at a fraction of the memory required for full-MDP analysis. We note that the optimal policy obtained via full-MDP model checking achieves a
stockout probability of approximately $3.14 \times 10^{-10}$ and an
inventory-full probability of exactly $0$, substantially lower than the
PPO policy's 2.9\% and 1.1\%, respectively. However, full-MDP synthesis
is only feasible up to a certain scale; as the environment becomes more
fine-grained (e.g., higher inventory capacities, finer shelf-life
discretizations, or additional state variables), the full MDP quickly
becomes intractable, and \emph{COOL-MC}'s induced-DTMC approach remains tractable longer~\cite{DBLP:conf/setta/GrossJJP22}.

Using the until operator, we query
$P_{=?}(\mathit{tinv} \geq 5 \;\mathcal{U}^{\leq 200}\; \textit{empty})$.
This measures
the probability that inventory remains at a level ($\geq 5$ units)
at every step \emph{until} a complete stockout occurs. The result is $0.015$: only $1.5\%$ of trajectories reach a
stockout while the inventory has never dropped below 5 units beforehand,
i.e.\ the system transitions directly from a safe level to complete
depletion in a single demand spike. Compared to the total stockout
probability of $2.9\%$, this implies that roughly half of all stockouts
are preceded by a gradual decline through a low-stock warning zone
($\mathit{tinv} < 5$), giving a human operator advance notice to
intervene, while the remaining half occur without such warning.

\subsubsection{Part B: Behavioral analysis}

To understand \emph{how} the trained policy manages the platelet
inventory, we perform four analyses on the induced DTMC: action reachability, feature pruning sensitivity, feature
importance timing, and action replacement counterfactuals.

\begin{figure}
    \centering
    \scalebox{0.5}{
        \input{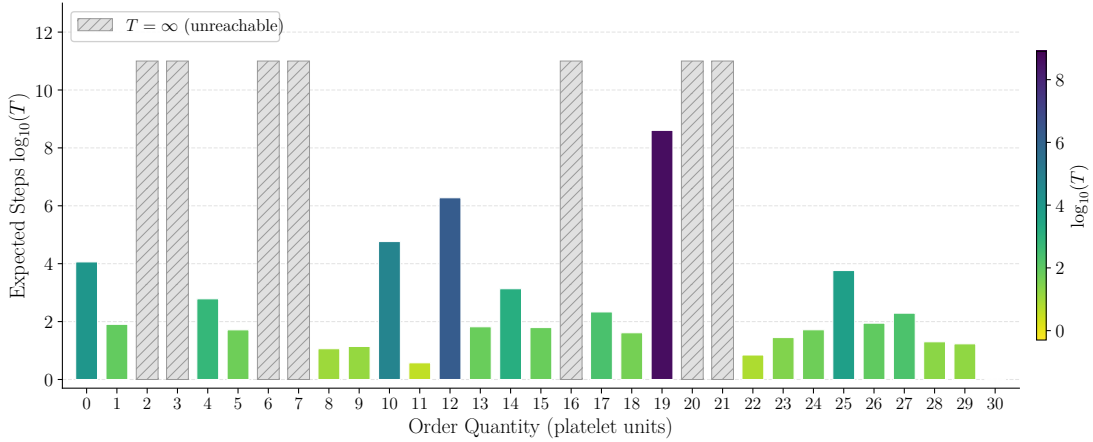}
    }
    \caption{Expected number of steps to the first occurrence of each ordering action under the trained PPO policy, computed via PCTL queries $T_{=?}(\,F\;\texttt{"pr*}\texttt{"}\,)$ on the induced DTMC. The $y$-axis shows $\log_{10}(T)$; hatched bars indicate actions that are never selected ($T = \infty$). Actions with low $T$ dominate the policy's behaviour, while extreme order quantities are either rarely reached or entirely unreachable.}
    \label{fig:action_expected_steps}
\end{figure}

\paragraph{Action reachability.}
Figure~\ref{fig:action_expected_steps} reports the expected number of
steps until each ordering action is first selected under the trained
policy, computed via PCTL queries $T_{=?}(\,F\;\texttt{"pr}X\texttt{"}\,)$
on the induced DTMC.

The results reveal that, apart from seven actions
that are never chosen ($T = \infty$) and a few that are reached only after a large number of steps (e.g.,
$\texttt{pr}_{19}$ with $T \approx 4 \times 10^8$), the majority of order quantities are selected within a relatively short time period, indicating a diverse replenishment strategy.

\begin{figure}
    \centering
    \scalebox{0.5}{
        \input{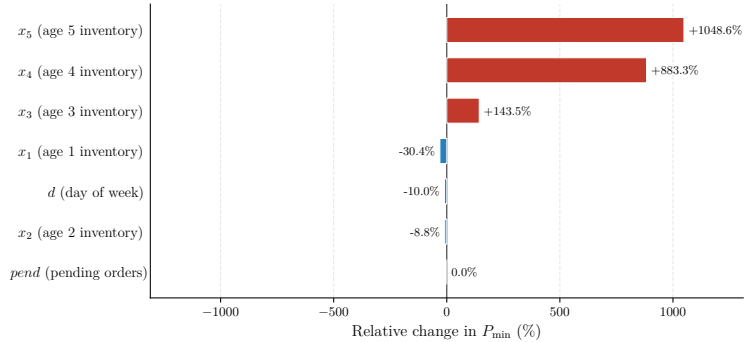}
    }
    \caption{Relative change in the probability of reaching an empty inventory state, $P_{=?}(\lozenge^{\leq 200}\;\texttt{"empty"})$, when each feature is individually pruned from the learned PPO policy. Positive values (red) indicate that removing the feature \emph{increases} the probability of a stockout, revealing features critical for the avoidance of inventory depletion. Notably, the freshest inventory levels $x_4$ and $x_5$ are by far the most influential, suggesting that the policy heavily relies on fresh stock information to prevent shortages.}

    \label{fig:pruning_empty}
\end{figure}

\begin{figure}
    \centering
    \scalebox{0.5}{
        \input{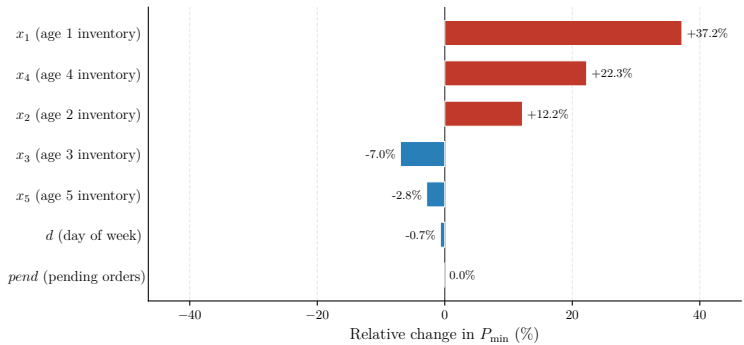}
    }
    \caption{Relative change in the probability of reaching a full inventory state, $P_{=?}(\lozenge^{\leq 200}\;\texttt{"full"})$, when each feature is individually pruned from the learned PPO policy. Positive values (red) indicate that removing the feature \emph{increases} the probability of overstocking. The oldest inventory level $x_1$ has the strongest influence, followed by $x_4$ and $x_2$, indicating that the policy primarily leverages old stock information to regulate ordering and avoid surplus. In contrast, pending orders ($pend$) and day of week ($d$) have a negligible impact on the overstocking probability.}

    \label{fig:pruning_full}
\end{figure}

\paragraph{Feature pruning sensitivity.}
To quantify each feature's contribution to the policy's safety
properties, we individually zero out each feature's weights in the
network's first layer and re-evaluate the induced DTMC.
Figure~\ref{fig:pruning_empty} shows the effect on the stockout
probability $P\!(\lozenge^{\leq 200}\;\textit{empty})$.
Removing the freshest inventory levels $x_5$ and $x_4$ (i.e., setting these features to a constant input of zero) increases the
stockout probability by over $1{,}048\%$ and $883\%$, respectively. Conversely, removing the oldest stocks ($x_1$, $x_2$) and $d$
slightly \emph{decreases} the stockout probability (by up to $30\%$),
suggesting that these features introduce minor decision noise with
respect to this particular safety property.
Removing pending orders has no effect on performance.

Figure~\ref{fig:pruning_full} shows a complementary picture for the
inventory-full probability
$P\!(\lozenge^{\leq 200}\;\textit{full})$. Here, the oldest
inventory level $x_1$ is the most influential ($+37\%$), followed by
$x_4$ ($+22\%$) and $x_2$ ($+12\%$), indicating that the policy
relies on old and mid-aged stock information to regulate ordering and
prevent overstocking.
Conversely, removing $x_3$ or $x_5$ slightly
\emph{decreases} the inventory-full probability (by $7\%$ and $3\%$,
respectively), again pointing to minor decision noise from these
features in this context.
The pending orders feature
$pend$ has zero impact on either the stockout or the inventory-full probability, suggesting that the policy has learned to infer sufficient ordering information from the inventory state alone.

\begin{figure}
    \centering
    \scalebox{0.5}{
    \input{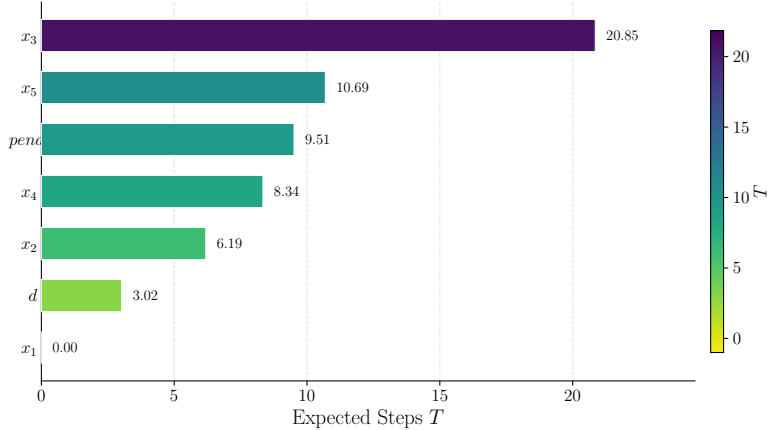}
    }
    \caption{Expected number of steps $T$ until the most important feature
    value is first encountered under the trained policy, computed via
    $T_{=?}\!\left(\,F\;\texttt{"imp\_}f\texttt{"}\,\right)$ for each
    feature~$f$. Higher values indicate that the feature's critical region
    is reached less frequently.
    Feature~$x_1$ yields
    $T = 0$, meaning its most important value is immediately present in
    the initial state. Feature~$x_3$ dominates with
    $T = 20.85$, indicating its value emerges only after extended
    system evolution.}
    \label{fig:feature_importance_T}
\end{figure}

\paragraph{Feature importance timing.}
Figure~\ref{fig:feature_importance_T} complements the pruning analysis
by revealing \emph{when} each feature's most important value is first
encountered during system evolution. Feature~$x_1$ has $T = 0$, meaning
its critical value is present in the initial state, consistent with
its role in regulating early ordering decisions. In contrast, $x_3$
dominates with $T = 20.85$ expected steps, indicating that its most
influential value emerges only after the inventory system has evolved
over multiple replenishment cycles. Features $x_5$ and
$pend$ occupy intermediate positions ($T \approx 10.7$ and
$T \approx 9.5$, respectively), while the day-of-week feature $d$ is
encountered early ($T = 3.02$).

\paragraph{Action replacement counterfactual.}
To probe the role of individual actions, we perform a
counterfactual analysis using \emph{COOL-MC}'s action replacement mechanism.
Specifically, we replace every occurrence of $\texttt{pr}_{14}$
(ordering 14~units), which accounts for 11.8\% of the reachable states,
with $\texttt{pr}_6$ (ordering 6~units) and reconstruct the induced
DTMC. Despite more than halving the order quantity in a substantial
fraction of states, the effect on both safety properties is negligible: the stockout probability shifts from $0.029$ to $0.0287$ and the
inventory-full probability from $0.0105$ to $0.0103$. This near-invariance
suggests that the policy deploys $\texttt{pr}_{14}$ in states where the system has sufficient inventory buffer to absorb a reduced order without compromising safety.

\section{Discussion}
This section discusses the capabilities that \emph{COOL-MC} demonstrates in this case study and reflects on the implications of our findings for platelet inventory management.

\subsection{From Aggregate Cost to Structural Understanding}
Standard RL evaluation of inventory management policies reduces performance to a single scalar: expected cost per period, combining shortage and outdating penalties.
This aggregate metric confirms whether a policy is cost-effective but does not reveal \emph{how} it achieves that outcome, such as how it balances the asymmetric costs of shortages ($C_S = 5$) and full inventories ($C_O = 1$).

\emph{COOL-MC} moves beyond this aggregate view by constructing the
induced DTMC for PCTL queries, feature pruning, and counterfactual
analysis. These reveal, for instance, that six order quantities are
never selected, that the freshest inventory features $x_4$/$x_5$ are critical for
stockout avoidance, while the oldest stock $x_1$ governs the inventory-full
prevention, and that replacing $\texttt{pr}_{14}$ with
$\texttt{pr}_6$ in $11.8\%$ of states leaves both safety probabilities
unchanged, transforming a single reward number into a structured
characterization of the policy's behaviour.

\subsection{COOL-MC as a Pre-Deployment Analysis Tool for Inventory Policies}
For blood bank operations, \emph{COOL-MC}'s analysis pipeline could support several practical workflows.
First, before deploying a new ordering policy, managers could verify that it meets safety requirements.
Second, when evaluating policy changes, the feature pruning profile could confirm that the new policy attends to the relevant inventory features.
Third, MDP parameter sweeps could assess policy robustness under anticipated demand changes (e.g., seasonal variation, pandemic-related disruptions) without requiring retraining for each scenario.

\subsection{Additional COOL-MC Capabilities}\label{sec:extra}
In this work, we focused on PPO training, PCTL verification on induced DTMCs, feature pruning, temporal feature-importance permutation ranking, and action labeling with parameter sensitivity.
However, \emph{COOL-MC} offers additional capabilities that could further enrich the analysis of platelet inventory policies.

\emph{COOL-MC}'s feature abstraction and feature remapping
capabilities~\cite{DBLP:conf/setta/GrossJJP22} allow analysis of policy
behaviour when state features are coarsened. In an inventory context, this could aggregate adjacent stock levels to simulate different measurement precision levels, quantifying how much safety degrades when the blood bank operates with coarser inventory tracking.

\emph{COOL-MC} also supports robustness verification through adversarial state-observation perturbations~\cite{DBLP:conf/icaart/GrossS0023,DBLP:conf/aips/GrossS0023}.
By perturbing the inventory observation fed to the policy and re-verifying the induced DTMC, one can assess whether small counting errors or measurement inaccuracies in the recorded stock levels cause the policy to make substantially different ordering decisions.
In a blood bank setting, this could reveal whether the policy is robust to minor inventory discrepancies arising from manual counting or delayed database updates.

\section{Conclusion}
This paper demonstrated the application of \emph{COOL-MC} to the platelet inventory management MDP inspired by Haijema et al., providing the first formal probabilistic verification and explanation of an RL-trained ordering policy for this domain.

Several directions for future work emerge from this study. These include applying \emph{COOL-MC} to other healthcare MDPs~\cite{valizadeh2026platelet,soares2020optimisation} and integrating domain-expert knowledge into the analysis.
Furthermore, leveraging policy-agnostic test suite methods, such as MPTCS~\cite{18326}, to identify critical initial states can focus bounded reachability verification on the most challenging regions of the state space.
\bibliographystyle{plain}
\bibliography{refs}

\end{document}